\documentclass[twocolumn]{article}
\usepackage{arxiv}
\usepackage[utf8]{inputenc} 
\usepackage[T1]{fontenc}    
\usepackage{hyperref}       
\usepackage{url}            
\usepackage{booktabs}       
\usepackage{amsfonts}       
\usepackage{nicefrac}       
\usepackage{microtype}      
\usepackage{graphicx}
\usepackage[numbers]{natbib}
\usepackage{doi}
\usepackage{subcaption}
\usepackage{algorithm}
\usepackage{algorithmicx}
\usepackage{algpseudocode}
\usepackage{appendix}
\usepackage{array}
\usepackage{soul}

\usepackage{multirow}%
\usepackage{amsmath,amssymb,amsfonts}%
\usepackage{amsthm}%
\usepackage{mathrsfs}%
\usepackage{textcomp}%
\usepackage{manyfoot}%
\usepackage{listings}%
\usepackage{orcidlink}
\usepackage{physics}
\usepackage{tikz}
\usetikzlibrary{calc}
\usepackage{xurl}
\usepackage{lmodern}
\usepackage{anyfontsize}
\usepackage{pgfplots}
 \pgfplotsset{compat=1.18}
\usepackage{caption}
\usepackage{verbatim}

\title{Physics Inspired Neural Networks Enable Efficient Graph Coloring}

\author{\href{https://orcid.org/0009-0007-9632-9204}    {\hspace{1mm}Lorenzo~Colantonio}\\
	Department of Physics\\
	Sapienza University of Rome\\
	  Piazzale Aldo Moro 5, Roma - 00185, Italy \\
	\texttt{lorenzo.colantonio@uniroma1.it} \\
	\And
	\href{https://orcid.org/0000-0003-2849-0075}{\hspace{1mm}Andrea~Cacioppo} \\
	Department of Physics\\
	Sapienza University of Rome\\
	  Piazzale Aldo Moro 5, Roma - 00185, Italy \\
	\texttt{andrea.cacioppo@uniroma1.it} \\
 \And
	\href{https://orcid.org/0009-0001-3248-8528}
 {\hspace{1mm}Federico~Scarpati} \\
	Department of Physics\\
	Sapienza University of Rome\\
	  Piazzale Aldo Moro 5, Roma - 00185, Italy \\
	\texttt{federico.scarpati@gmail.com} \\
 \And
	\href{https://orcid.org/0000-0002-6277-359X}{\hspace{1mm}Maria~Chiara~Angelini} \\
	Department of Physics\\
	Sapienza University of Rome\\
	  Piazzale Aldo Moro 5, Roma - 00185, Italy \\
	\texttt{mariachiara.angelini@uniroma1.it} \\
 \And
	\href{https://orcid.org/0000-0003-4970-7376}{\hspace{1mm}Federico~Ricci-Tersenghi} \\
	Department of Physics\\
	Sapienza University of Rome\\
	  Piazzale Aldo Moro 5, Roma - 00185, Italy \\
	\texttt{federico.ricci@uniroma1.it} \\
 \And
	\href{https://orcid.org/0000-0001-9192-3537}{\hspace{1mm}Stefano~Giagu} \\
	Department of Physics\\
	Sapienza University of Rome\\
	  Piazzale Aldo Moro 5, Roma - 00185, Italy \\
	\texttt{stefano.giagu@uniroma1.it} \\
}

\date{}

\renewcommand{\headeright}{}
\renewcommand{\undertitle}{}
\renewcommand{\shorttitle}{Physics Inspired Neural Networks Enable Efficient Graph Coloring}

\hypersetup{
pdftitle={Physics Inspired
Neural Networks Enable Efficient Graph Coloring},
pdfsubject={Machine Learning, Statistical Physics, Optimization and Control},
pdfauthor={Lorenzo~Colantonio, Andrea~Cacioppo, Federico~Scarpati, Maria~Chiara~Angelini, Federico~Ricci-Tersenghi, Stefano~Giagu},
pdfkeywords={Graph coloring, Physics-inspired Machine Learning, Graph neural networks, Optimization Problem, Potts model},
}

\begin{document}

\twocolumn[
\maketitle
\begin{abstract}
    \vspace{0.5cm}
    Combinatorial optimization problems near algorithmic phase transitions represent a fundamental challenge for both classical algorithms and machine learning approaches. Among them, graph coloring is a prototypical constraint satisfaction problem characterized by phase transitions that fragment the solution space into distinct clusters, making the search for valid solutions challenging. Here we introduce a physics-inspired neural framework that learns to solve large-scale graph coloring instances by combining graph neural networks with statistical-mechanics principles. Our approach integrates a planting-based supervised signal, symmetry-breaking regularization, and iterative noise-annealed neural dynamics to navigate complex solution landscapes. When the number of iterations scales quadratically with graph size, the learned solver successfully tackles increasingly challenging instances, even in regimes where finding valid solutions becomes particularly difficult, demonstrating its ability to extract meaningful information from the graph structure. The model generalizes from small training graphs to instances up to 20 times larger, demonstrating that neural architectures can learn scalable algorithmic strategies that remain effective in hard connectivity regions. These results establish a general framework for learning neural solvers that operate near fundamental phase boundaries in combinatorial optimization and inference.
    \vspace{0.5cm}
\end{abstract}

]

\clearpage

\section{Introduction}\label{sec:introduction}

The graph coloring problem (GCP) is a paradigmatic constrained optimization problem that lies at the core of combinatorial complexity theory. It consists of assigning one among $q$ possible colors to each vertex of an undirected graph, so that no two adjacent vertices share the same color. 
The problem of deciding whether a graph can be fairly colored with $q$ colors is NP-complete \cite{karp2010reducibility}, while the optimization of a graph's coloring to minimize conflicts is NP-hard \cite{garey1979computers}.
Beyond its theoretical relevance, graph coloring arises in a wide range of practical applications, including scheduling \cite{lewis2021guide}, timetabling \cite{ahmed2012applications} and even puzzle solving such as Sudoku \cite{tosuni2015graph}, moreover it can be reduced in polynomial time to all other NP-hard problems \cite{karp2010reducibility} as the travelling salesman problem, the clique problem or k-SAT \cite{garey1979computers}.
The intrinsic difficulty of the GCP is deeply connected to the widely held belief that $P \neq NP$  \cite{garey1979computers} \cite{fortnow2009status}, making the GCP both difficult and useful to solve. 
In large random graphs, the structure of the solution space undergoes sharp phase transitions as the average connectivity varies. These transitions dramatically affect the performance of algorithms, creating regimes in which solutions exist but are algorithmically hard to find. Understanding and overcoming these barriers remains a central challenge in combinatorial optimization.
Classical approaches to graph coloring typically rely on local search in configuration space \cite{galinier2006survey}. Markov Chain Monte Carlo (MCMC) methods such as Simulated Annealing \cite{chams1987some} \cite{johnson1991optimization} \cite{angelini2023limits}, TabuCol \cite{hertz1987using}, and evolutionary algorithms \cite{galinier1999hybrid, malaguti2008metaheuristic, mostafaie2020systematic}, can achieve strong performance, but may require running times that grow exponentially with graph size in hard regimes. Greedy heuristics such as Greedy coloring \cite{matula1983smallest} or DSatur \cite{brelaz1979new}, avoid the problem of time scaling but provide sub-optimal solutions. In recent years, graph neural networks (GNNs) have emerged as a promising alternative, leveraging the connectivity structure to process graph-structured data. Several works have applied GNNs to graph coloring, using reinforcement learning, direct classification of colorability, or physics-inspired energy minimization strategies. While these approaches demonstrate encouraging results, their performance often degrades near critical connectivity thresholds, precisely where the problem becomes structurally hardest.

In this work, we introduce a physics-inspired neural framework that explicitly integrates statistical-mechanics insights into the training and inference dynamics of a GNN-based solver. 
By scaling the number of inference iterations with the problem size,
we demonstrate that the learned solver remains  effective in challenging regimes where the solution space becomes increasingly difficult. Furthermore, it can recover the planted solution \cite{krzakala2009hiding}, a valid coloring embedded during graph generation, even when the underlying structure is difficult to detect.. Moreover, the model generalizes from training on relatively small graphs to solving instances up to 20 times larger, highlighting the potential of neural architectures to learn scalable algorithmic strategies rather than memorizing instance-specific patterns.
This paper is structured as follows. In section~\ref{sec:related_work}, we provide a high-level overview of existing methods based on GNNs. In Sec.~\ref{sec:background} we address the GCP from a statistical mechanics point of view, highlighting its connections with the Potts model. We describe the two types of graphs used in this work, namely Erdős–Rényi and planted graphs, describing the procedure for generating them. In Sec.~\ref{sec:methods} we detail the dataset used in this work, we present a GNN-based architecture, introduce an effective supervised training strategy, and describe the coloring procedure. In Sec.~\ref{sec:results} we presents the performances of our method in terms of energy and  at different connectivities, comparing with the best state-of-the-art graph coloring algorithms.
 
\section{Related Work}\label{sec:related_work}

There is a vast and diverse list of algorithms used to solve NP-hard problems. In this section, we focus on algorithms that primarily use GNNs for solving the GCP.

In \cite{huang2019coloring}, improved heuristics for graph coloring are sought using reinforcement learning, leveraging a deep neural network architecture that has access to the entire graph structure. In \cite{lemos2019graph}, a GNN is used to determine whether a graph can be colored using $q$ colors or not. Additionally, it is shown how node embeddings can be used to assign coloring to the graph. In \cite{li2022rethinking}, the performances of aggregation-combine GNNs are studied. The work highlights the features that limit the expressive power of these models in solving node-classification problems and proposes solutions to improve performances.

In \cite{schuetz2022graph}, the authors use the architecture of a physics-inspired GNN (PI-GNN) for graph coloring. Specifically, the neural network is optimized in order to minimize the Potts energy separately for each graph instance. This approach has been adopted in several subsequent works. In \cite{wang2023graph}, an alternative to PI-GNN is proposed, introducing negative-message-passing, achieving numerical improvements. Additionally, an entropic term is introduced in the loss function to accelerate convergence during the model training. Finally, in \cite{zhang2024using}, a graph isomorphism network model, using a physics-inspired approach, is proposed. The output of the neural network, is post-processed using TabuCol to reduce the number of conflicts.
Our work builds upon the same continuous Potts formulation and a related message-passing architecture, but differs fundamentally in the learning and inference framework. Rather than optimizing the model for each graph, we train a \emph{single} GNN on a distribution of graphs. The trained architecture is then used to solve unseen graphs through an iterative noise-annealed dynamics.
Our approach combines three key ingredients: (i) a planting procedure that can produce large, hard instances with associated labels, suitable for supervised training of the network, while preserving, in relevant regimes, the same statistical properties as random graph ensembles (ii) a semi-supervised loss that blends a differentiable Potts energy with symmetry-breaking overlap terms; and (iii) an iterative noise-annealed inference procedure that enables the model to escape metastable configurations in clustered solution landscapes.

\section{Background}\label{sec:background}

In this section, we formally introduce the GCP and its connection to the Potts model, furthermore the graphs ensembles used in this work are presented and the algorithm used to generate them is described.

\subsection{Graph coloring}
The GCP is a very well-known constraint satisfaction problem (CSP). Formally, we consider an undirected graph ${\mathcal{G} = (\mathcal{V}, \mathcal{E})}$ where ${\mathcal{V} = \{1, ..., N\}}$ is the set of vertices and ${\mathcal{E} = \{(i,j)| i,j \in \mathcal{V}\}}$ is the set of edges. 
The objective is to assign an integer variable (or color) ${s_{\nu} \in \{1, ..., q\}}$ to each node $\nu \in \mathcal{V}$, such that nodes sharing the same edge have different colors, namely
\begin{equation}
 s_i\neq s_j, \  \forall (i,j)\in \mathcal{E} 
\end{equation}
If a graph admits such a configuration $\mathbf{s}$, the graph is said $q$-colorable. The smallest value of $q$ for which the graph is colorable is the chromatic number $\chi (\mathcal{G})$.

\subsection{Potts model}\label{par:potts}

Searching graph coloring solutions is equivalent to studying the energy minima of the anti-ferromagnetic Potts model. The Potts model is the generalization of an Ising model, where the spins can take $q$ different values \cite{wu1982potts}. The energy of the anti-ferromagnetic Potts model is given by
\begin{equation}
    H_\text{Potts}(\mathbf{s}) = \sum_{(i,j)\in \mathcal{E}} \delta_{s_i, s_j}\;,
\end{equation}
and counts the number of conflicting edges in the associated GCP.
For a given graph $\mathcal{G}$ in equilibrium at temperature $T$, the nodes configurations $\mathbf{s}$ follows the Gibbs-Boltzmann probability distribution
\begin{equation}
    P_{\mathcal{G}}(\mathbf{s})=\frac{e^{-H(\mathbf{s})/T}}{\mathcal{Z}}\;,
\end{equation}
where the partition function $\mathcal{Z}$ is the normalization factor.
In the limit $T \to 0$, the measure is dominated by configurations with the minimum energy $E_0$. If $E_0=0$, the graph is $q$-colorable. Sampling the $T=0$ distribution solves the GCP \cite{mezard2009information} and underlies algorithms such as simulated annealing (SA) \cite{kirkpatrick1983optimization} and belief propagation (BP)\cite{pearl2022reverend}.
Nevertheless, sampling the Gibbs distribution is a considerably stronger requirement than finding a single valid coloring. In practice, SA and BP-based strategies---such as BP with decimation or reinforcement---are both unable to sample beyond the so-called dynamical threshold \cite{ghio2024sampling}, and BP-based strategies fails even before it \cite{ricci2009cavity}. Our method therefore addresses only the latter: finding one valid coloring.

It is well known that the solution space of GCP for random graphs such as Random Regular Graphs (RRG) \cite{bollobas1998random} and Erdős–Rényi Graphs (ERG) \cite{erdds1959random} undergoes multiple phase transitions \cite{zdeborova2007phase}.  
\begin{figure*}[t]
    \centering
    \includegraphics[width=\textwidth]{phase_transitions.pdf}
    \caption{
Schematic illustration of the evolution of the solution-space landscape
with increasing average connectivity $c$. The red cross represent the
planted solution. At low connectivity, the planted solution belongs to a
large connected region of solutions. Above the dynamical transition $c_d$, the solution space fragments into an exponential number of clusters. At the condensation transition $c_c$, the solution measure concentrates onto a subexponential number of dominant clusters, one of which contains the planted solution. The satisfiability threshold $c_s$ marks the transition beyond which typical random instances become unsatisfiable. The Kesten--Stigum threshold $c_{\mathrm{KS}}$ marks the detectability of the planted structure. Blue regions denote solutions, whereas orange regions represent local minima.
}
    \label{fig:phase_transition}
\end{figure*}
The success of sampling and the time required to perform it strongly depend on the average connectivity of the graph under consideration and the number of colors $q$. In particular, for sufficiently small connectivity values $c<c_d(q)$, most solutions form a single connected cluster, where any two solutions differ by only a sub-extensive number of node variables. In this phase, finding solutions is particularly easy for most algorithms. In the $q>3$ case, at $c_d(q)$ a dynamic phase transition takes place, such that for $c>c_d(q)$ the solutions are shattered among exponentially many (in $N$) clusters; this structure of solutions generally slows down the dynamics of algorithms.
In the range $c_c(q)<c<c_s(q)$, the number of clusters becomes sub-exponential, making feasible colorings difficult to find using polynomial-time algorithms. Finally, in the region with high connectivity values $c>c_s(q)$, the probability that a random graph is $q$-colorable vanishes in the large $N$ limit. We stress that the above scenario holds only in the thermodynamic limit, while for finite-size graphs, there are exeption to this behavior. The $q\leq 3$ case is less significant, as the $c_d(q)$ and $c_c(q)$ critical thresholds coincide, resulting in a situation that is generally easier to tackle for algorithms.

\subsection{Planted Erdős–Rényi random graphs}

As we have discussed in Secs.~\ref{sec:introduction} and \ref{par:potts}, finding a perfect coloring for a generic graph can be very resource-expensive. To this end, in the present paragraph, we briefly discuss a simple and effective method that allows us to generate graphs for which a solution $\mathbf{s}^*$ is already known \cite{krzakala2009hiding}. 
The procedure is explained in Algorithm A.1 in the Supplementary Information.
Notably, this algorithm can generate fairly colored graphs at any connectivity, even beyond the satisfiability threshold ($c>c_s$).
It can be proved that planted random graphs are indistinguishable from random graphs as long as $c<c_c$ \cite{krzakala2009hiding}.
For this reason, in that region, the procedure is called quiet-planting.
When $c>c_c$, the entropy of the planted state (i.e., the set of solutions around the planted one) becomes larger than the entropy of all other random solutions, making the planted random graph distinguishable from a standard (non-planted) random graph.
However, the model that we introduce in Sec.~\ref{sec:methods} does not aim to sample graphs from the Gibbs distribution, but to find a solution that minimizes the energy.
Therefore, the solutions obtained by the planting procedure continue to be suitable for the purpose of training.

In the planted model, the planted solution can be detected only if the value of $c$ is large enough. At present, the best detection algorithm is Belief Propagation, which is successful for $c>c_\text{\tiny KS}$, where $c_\text{\tiny KS}$ is the Kesten-Stigum bound. Originally derived by Kesten and Stigum in the study of the asymptotic growth of multitype branching processes \cite{kesten1966additional}, the KS criterion was later connected to reconstruction problems on trees \cite{mossel2001reconstruction,mezard2006reconstruction}.
The KS threshold corresponds to the instability of the uninformative BP fixed point toward the planted solution and its role in inference problems has been discussed in detailed in the literature \cite{krzakala2009hiding, ricci2019typology}.
For the GCP the KS threshold marks the onset of detectability for polynomial-time algorithms.
In Fig.~\ref{fig:phase_transition} we report a visual representation of the solution space in each phase.

\section{Experiments}\label{sec:results}
In this section, we show the performance of our algorithm in coloring graphs.
Throughout this work, we focus on the representative case of $q=5$. As a matter of fact, the qualitative phase diagram is the same for all $q\ge5$, with the clustering, condensation, and satisfiability transitions occurring at different critical connectivities but exhibiting the same underlying behaviour. The situation is different for $q<5$: there, some transitions change order or collapse together \cite{zdeborova2007phase}, and the solution space is less fragmented, making the search problem substantially simpler \cite{Skenderi2026}. Our choice of $q=5$ is therefore motivated by the fact that it is the smallest number of colors for which the problem exhibits the full complexity of the hard regime.
We consider two configurations targeting distinct regions of the phase diagram: the Dynamical Transition Regime (DTR), which addresses random-graph coloring near the dynamical transition, and the Kesten–Stigum Regime (KSR), which focuses on planted inference near the Kesten–Stigum threshold.
\subsection{Models setup}
In the following, we provide details on the training datasets and architectures of the two models. Additional information on the model configuration, training procedure, and inference hyperparameters is provided in Table 1 and Table 2 of the Supplementary Information.

To train the DTR Model, we have used a dataset of $4000$ planted graphs for each connectivity in $[12.5, 12.7, 12.9, \dddot{},13.5, 13.7]$, and sizes $N=[1024,2048,4096]$ for a total of 84k graphs. This dataset has been proposed as a hard benchmark to test neural-network-based solvers \cite{Skenderi2026}.

To train the KSR Model, we have used a dataset of $10000$ planted graphs for each connectivity in $[15.5, 16.0, 16.5, 17.0]$, and size $N=1000$ for a total of 40k graphs.

For each model, $90\%$ of the dataset is used for training and the remaining $10\%$ for validation. The models are evaluated on independently generated graph instances.

In all models, we have used $q=5$ colors and an additional input feature to encode the node's degree. We have observed that this additional feature increases the model performance with a negligible computational cost. 

\subsection{Results}
In this section, we report the results obtained, showing the scaling properties of the models and comparing them to state-of-the-art algorithms.

To facilitate the interpretation of the results, we summarize the critical values for $q=5$: $c_d=12.837$, $c_c=13.23$ and $c_s=13.669$ for 5-coloring of random graphs \cite{zdeborova2007phase}, while $c_\text{\tiny KS}=16$ for detection of the planted configuration \cite{krzakala2009hiding}.
\begin{figure}[t]
    \centering
    \includegraphics[width=\columnwidth]{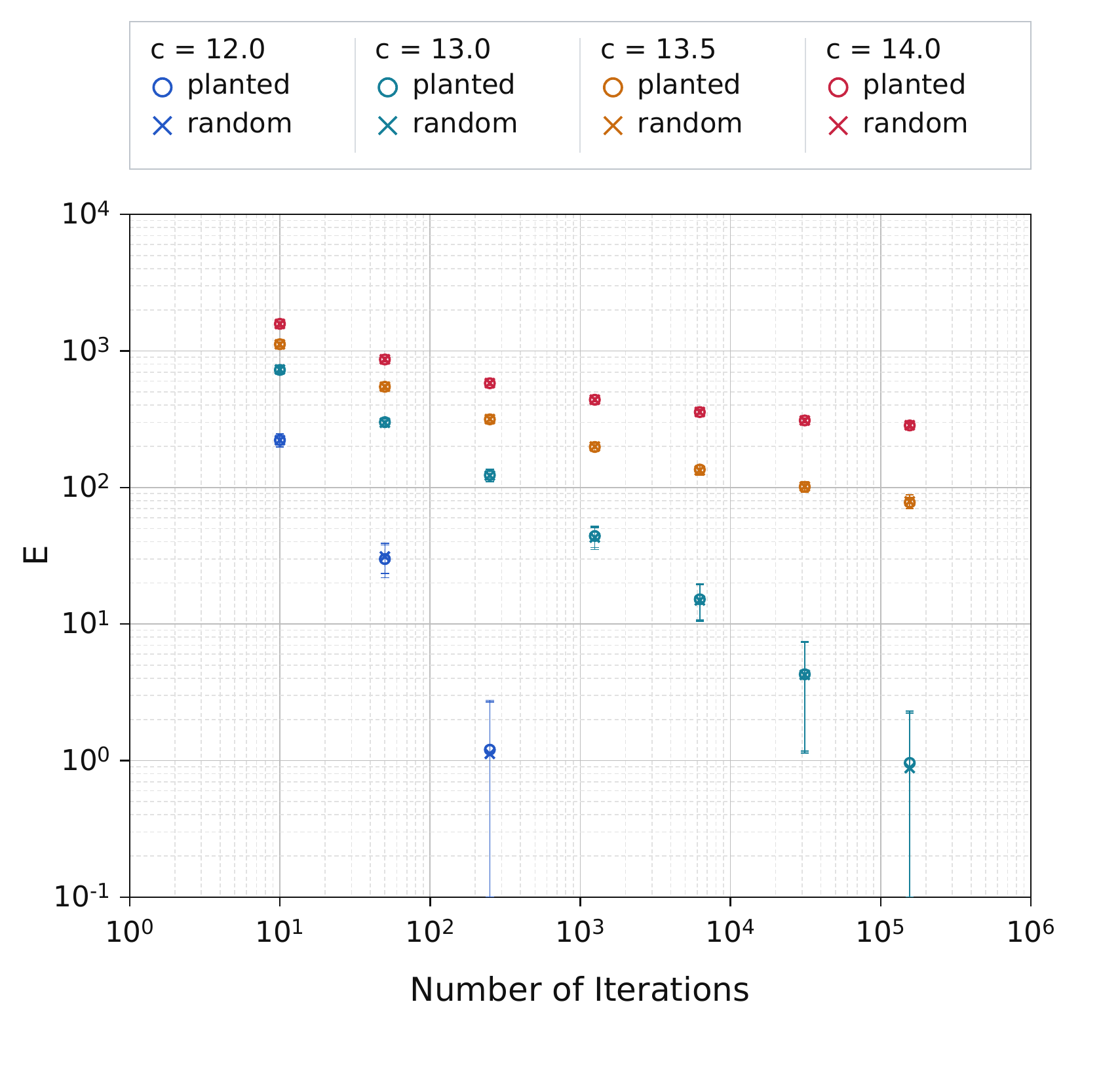}
    \caption{(DTR Model) Extensive energy as a function of the number of iterations for four representative connectivity values, $c = 12.0, 13.0, 13.5, 14.0$. Each data point is obtained by averaging results over 50 graphs with $N=$10k nodes. Error bars indicate one standard deviation across graph instances. Circles indicate values for planted graphs, while crosses indicate values for random graphs. Error bars indicate standard error of the mean
across graph instances.}
    \label{fig:modelA_scaling}
\end{figure}
\textbf{DTR Model}
Fig.~\ref{fig:modelA_scaling} shows the scaling with the number of iterations of the extensive energy $E$, which counts the number of unsatisfied edges. We plot data for 4 connectivities, corresponding to the 4 main phases of the solution space, separated by the critical values $c_d$, $c_c$, and $c_s$. As shown in Fig.~\ref{fig:modelA_scaling}, the scaling behaviour of the mean energy, which is similar for random and planted models in this range of connectivities, exhibits a clear dependence on the connectivity parameter $c$. For $c=12$, the energy decays faster than a power-law, indicating an efficient exploration of the energy landscape in the sparse regime. At $c=13.0$, the decay is approximately consistent with a power-law, suggesting a transition point where the algorithm still maintains an effective scaling with the number of iterations. It is worth stressing that for $c=13.0$ we are in a clustered phase, where other algorithms (e.g., equilibrium Monte Carlo samplings) would face serious limitations in converging to problem solutions. In contrast, for $c=13.5$ and $c=14.0$, the energy decreases significantly more slowly than a power-law, reflecting a pronounced slowdown in convergence.
\begin{figure}[t]
    \centering
    \includegraphics[width=\columnwidth]{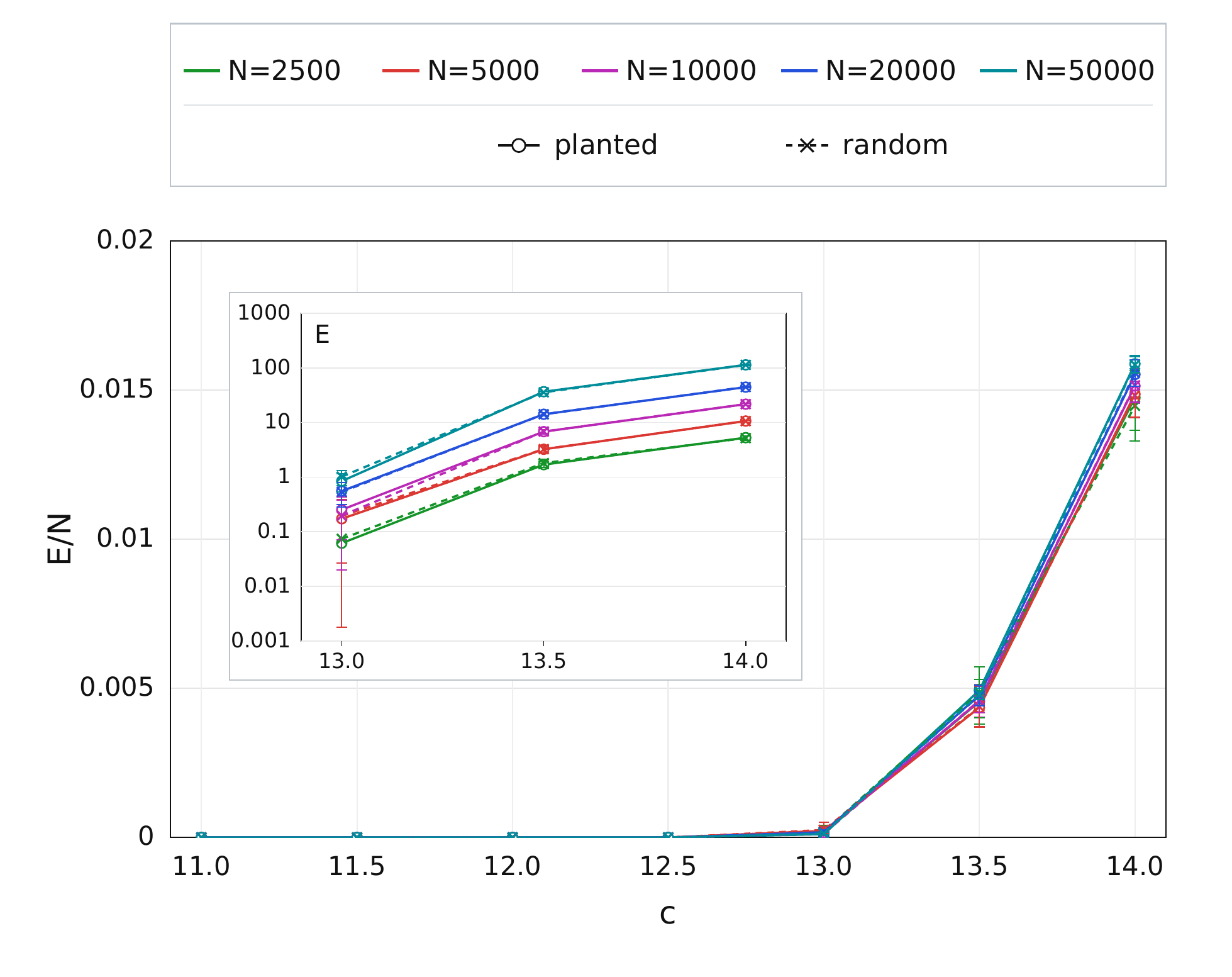}
    \caption{(DTR Model) \textbf{Main panel} Mean intensive energy as a function of connectivity, for both random and planted graphs, and a number of nodes varying from $N=2500$ to $N=50$k. Each data point is obtained by running the algorithm for $N_\text{iter}=200$k iterations and averaging the energy over 50 graphs. \textbf{Inset} The mean extensive energy grows with the size at each value of $c$ and does not allow to determine the algorithmic threshold in the large $N$ limit. Error bars indicate the standard error of the mean.}
    \label{fig:EvsC_modelA}
\end{figure}
The main panel of Fig.~\ref{fig:EvsC_modelA} shows the intensive energy $e=E/N$ as a function of the connectivity. It is worth noticing that data are very similar, for both random and planted graphs, in the whole range of sizes $3000\le N \le 50000$. Thus, we assume to have reached the large $N$ limit, and we can make the claim that the mean energy $e$ becomes non-zero close to the dynamical critical point $c_d$, and the model is not able to find solutions for larger values of $c$.

In the inset of Fig.~\ref{fig:EvsC_modelA}, we report the same data for the mean extensive energy $E$ on a logarithmic scale to make evident that, even in the region where $e$ looks very small, the number of violated constraints $E$ is actually growing with the graph size $N$.
Under these conditions, it is not possible to get any meaningful estimate for the algorithmic threshold $c_\text{alg}$. Indeed, for $c_\text{alg}$, in the large $N$ limit, the algorithm should be able to find a solution with high probability, and this implies that the mean extensive energy $E$ should go to zero.

The observation that an algorithmic threshold can not be estimated from the data shown in Fig.~\ref{fig:EvsC_modelA} suggests that the number of iterations should not be kept constant, but scaled with the problem size $N$ to properly identify $c_\text{alg}$. This has already been observed in the behaviour of other algorithms, like simulated annealing \cite{angelini2025algorithmic}, solving the same graph coloring problem.

\begin{figure}[t]
    \centering
    \includegraphics[width=\columnwidth]{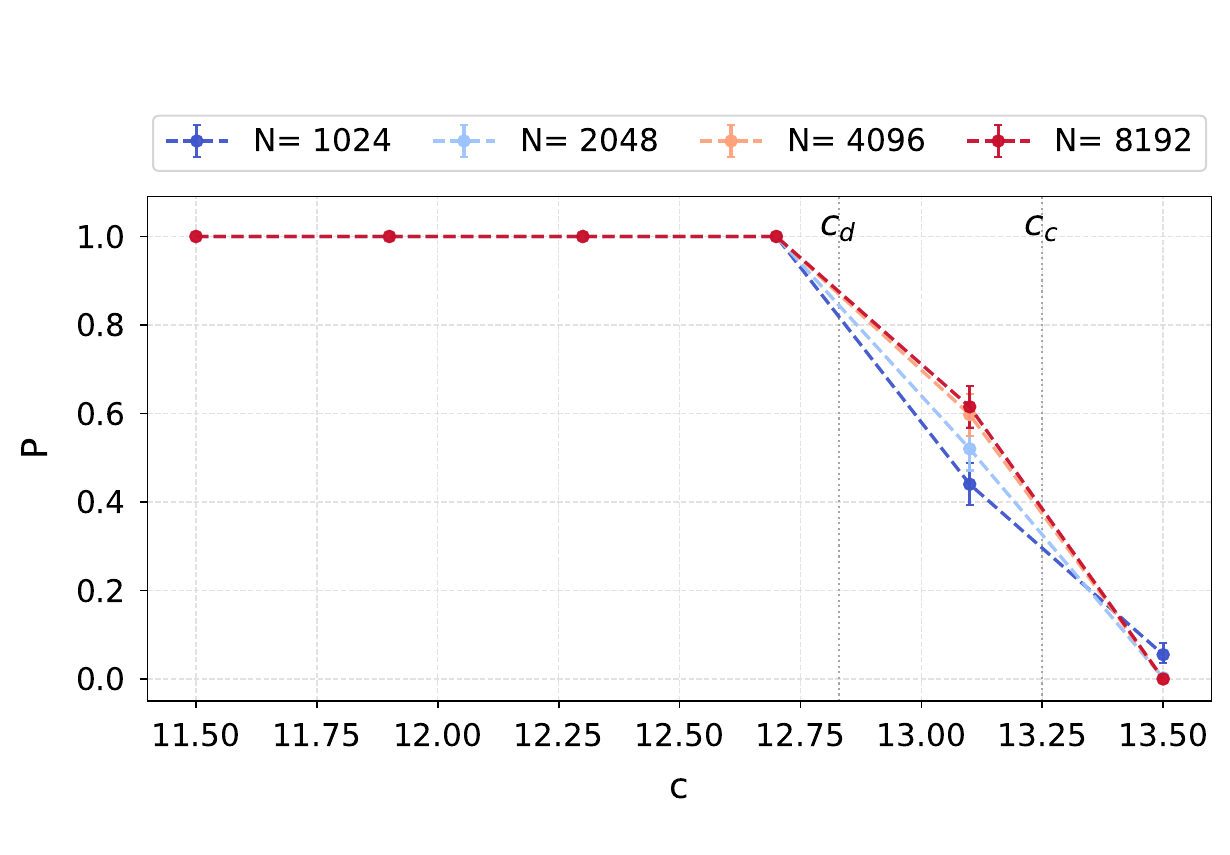}
    \caption{(DTR Model) Probability of finding a solution as a function of the graph connectivity $c$, for sizes $N\in[1024, 2048, 4096, 8192]$ and a number of iterations scaling as $N_\text{iter} = 0.04\,N^2$. Each point is the average over 400 graphs. Error bars indicate 95\% Wilson binomial confidence intervals.}
    \label{fig:sizecomparison_modelb}
\end{figure}

\begin{figure}[t]
    \centering
    \includegraphics[width=\columnwidth]{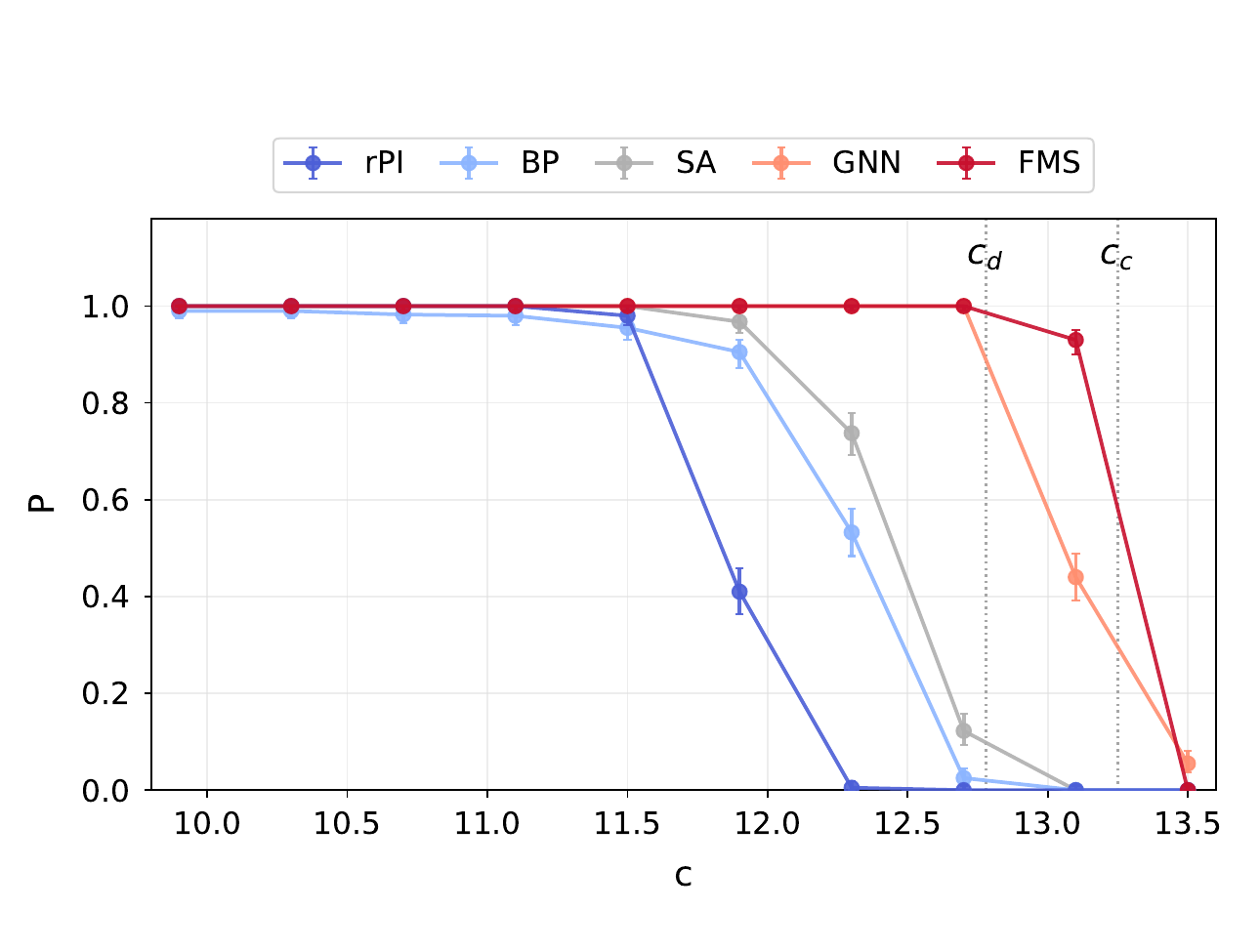}
    \caption{(DTR Model) Comparison of the probability of finding a solution as a function of the graph connectivity $c$ for different solving algorithms. Values are obtained by averaging over $400$ graphs of $N=1024$ nodes. The number of iterations $N_\text{iter}$ is chosen to ensure the same wall-clock execution time across all algorithms. The tested algorithms are a recurrent Physics-Inspired GNN (rPI) \cite{schuetz2022graph}, Belief Propagation (BP) \cite{Skenderi2026}, Simulated Annealing (SA) \cite{angelini2025algorithmic}, our graph neural network (GNN), and Focused Metropolis Search (FMS) \cite{Seitz_2005}.  Error bars indicate 95\% Wilson binomial confidence intervals.}
    \label{fig:algorithmcomparison_modelb}
\end{figure}
In Fig.~\ref{fig:sizecomparison_modelb} we plot the probability of finding a solution using the DTR Model run for a number of iterations scaling quadratically with the graph size, $N_\text{iter} = 0.04\,N^2$. The curves become sharper as the problem size $N$ increases and cross very close to $c_d$ (marked with a vertical dashed line). Thus we conclude that our solver based on a GNN run for a number of iterations scaling quadratically with the problem size has an algorithmic threshold very close to the dynamical transition, $c_\text{alg} \simeq c_d$.

Fig.~\ref{fig:algorithmcomparison_modelb} shows the probability of finding a solution in random graphs of size $N=1024$ for several smart algorithms, which are known to be the best available at present to solve the random graph coloring problem.
These algorithms have been already compared in recent studies \cite{angelini2025algorithmic,Skenderi2026}.
We notice that our GNN-based algorithm is performing very well, ranking second just after Focused Metropolis Search (FMS), which is known to have impressive performances in this class of hard optimization problems \cite{angelini2025algorithmic}.
Surpassing algorithms considered highly competitive in the field (Simulated Annealing, Belief Propagation, the Physics-Inspired GNN introduced in Ref.~\cite{schuetz2022graph}) is a major achievement of our algorithm.
To further assess the generalization capabilities of the proposed framework beyond the graph distribution used during training, we evaluated the trained model, on out-of-distribution instances drawn from standard graph-coloring benchmarks with chromatic number $\chi=5$. 
As shown in Table 3 in the Supplementary Information, the proposed algorithm successfully solves all the benchmark graphs considered. These results should not be interpreted as evidence of universal generalization across graph distributions. Rather, they show that the learned coloring strategy can transfer to structurally different instances.

\textbf{KSR Model}
While for $c<c_c$ random graphs and planted graphs are statistically indistinguishable, for $c>c_c$ the planted configuration is detectable.

In this regime, coloring of planted graphs is a prototypical hard inference problem,
with the planted coloring as the signal and the graph connectivity c as the signal-to-noise ratio.

Bayes optimal algorithms (like Belief Propagation \cite{krzakala2009hiding}) can detect the planted coloring with $q=5$ colors when $c>c_\text{\tiny KS}=16$, and this is considered the best achievable in polynomial time.
Indeed, most of the spectral algorithms show poorer performances \cite{krzakala2013spectral}.
Recently, the behavior of Simulated Annealing (SA) to solve this inference problem has been explored in detail: the standard version of SA has a larger algorithmic threshold ($c_\text{\tiny KS}^\text{\tiny SA}\simeq 18$), but an improved version with multiple interacting copies, replicated SA (rSA), exhibits an empirical detection threshold close to the Kesten–Stigum threshold ($c_\text{\tiny KS}^\text{\tiny rSA}\simeq 16$).

\begin{figure}[t]
    \centering
    \includegraphics[width=\columnwidth]{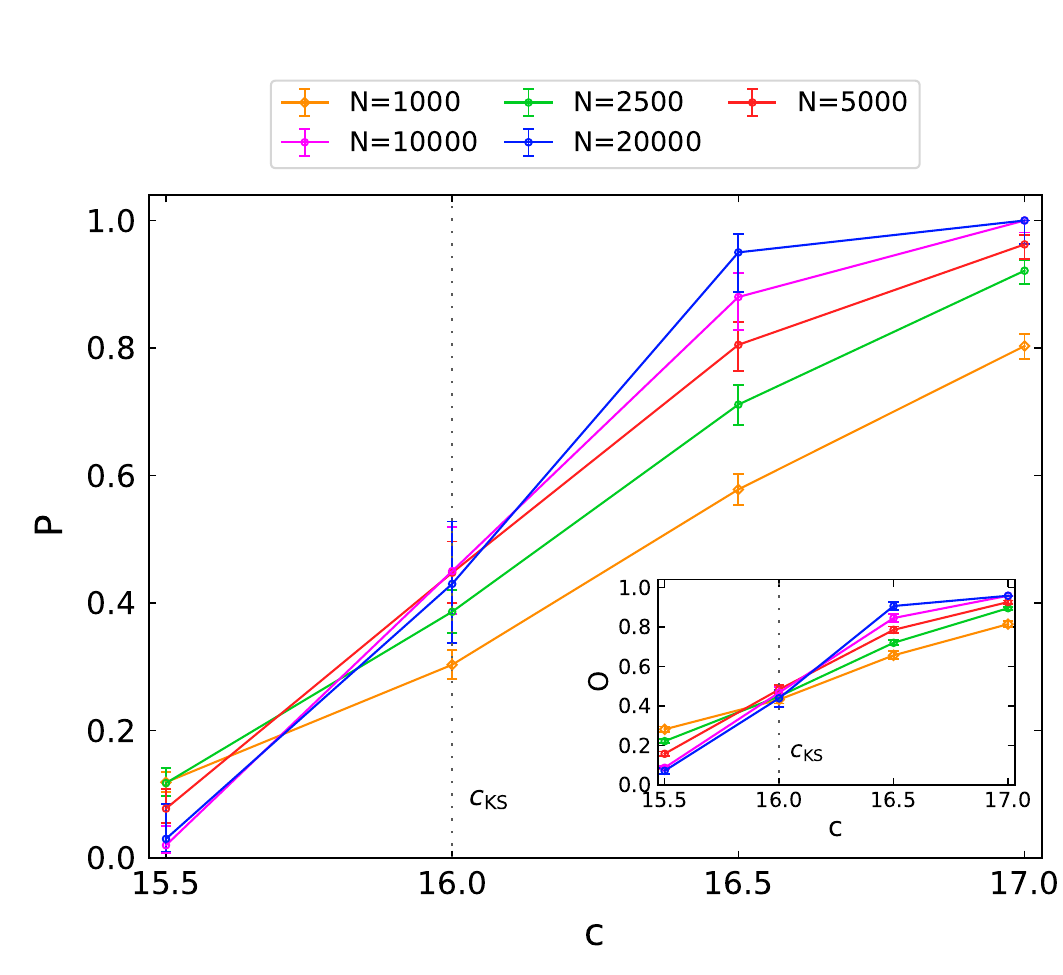}
    \caption{(KSR Model) \textbf{Main panel} Probability of finding a solution as a function of the graph connectivity $c$, for sizes $1000\le N\le 20000$ and a number of iterations scaling as $N_\text{iter} = 10^{-4}\,N^2$. Each point is the average over $2\,10^6/N$ graphs. \textbf{Inset} The mean overlap with the planted coloring of the configuration output by the algorithm. Error bars indicate 95\% Wilson binomial confidence intervals.}
    \label{fig:modelc}
\end{figure}

In the main panel of Fig.~\ref{fig:modelc} we plot the probability of finding a proper coloring of the graph as a function of the connectivity $c$, when the algorithm is run for a number of iteration growing quadratically with the problem size, $N_\text{iter} = 10^{-4}\,N^2$. For the range of graphs we have explored, $1000\le N\le 20000$, the curves show a weak size dependence and they sharply increase sharply near the Kesten–Stigum threshold $c_\text{\tiny KS}=16$.
Moreover the inset show the mean overlap with the planted coloring, and the similarity with the data presented in the main panel confirm that the solution found is (or is very close to) the planted one.
These finite-size results provide numerical evidence that the learned dynamics can recover configurations strongly correlated with the planted coloring at connectivities close to the Kesten–Stigum threshold.
This is once more an impressive result, making our GNN-based solver competitive with the state-of-the-art algorithms.

\section{Discussion}\label{sec:conclusions}

We have introduced an algorithm that leverages physics-inspired graph neural networks to solve efficiently the graph coloring problem, which provides hard instances of both optimization and inference problems depending on the random graph connectivity. 
Our algorithm clearly outperforms any machine-learning–based method previously available (see the recent comparative study in Ref.~\cite{Skenderi2026}) and it is competitive with other known algorithms: it ranks second, behind Focused Metropolis Search (an impressive local search algorithm) in the optimization problem, and it exhibits a detection transition close to the Kesten–Stigum threshold in the planted inference experiments.

A key conceptual contribution of this work is the integration of statistical-mechanics principles into the learning process. Impressive performances have been achieved implementing an approach that uses planted solutions to introduce supervised signals without sacrificing the statistical structure of hard random ensembles, while symmetry breaking and noise-annealed iteration enable controlled navigation of clustered energy landscapes, avoiding the common problem of over-smoothing in GNNs. It is also very important to stress that the reported performance near the phase-transition thresholds is obtained using a heuristic inference schedule in which the number of iterations scales quadratically with graph size, inspired by a recent work \cite{angelini2025algorithmic}.

Our results strongly suggest that further research in this direction is not only promising but necessary in order to push the current limits of machine learning methods used in high-dimensional combinatorial optimization and inference.  Our study shows that, for certain connectivity regimes, the performance of the algorithm does not significantly vary across different graph sizes. 
The ability of the trained model to generalize from relatively small graphs to instances up to 20 times larger highlights another important perspective. Instead of learning instance-specific mappings, the network appears to internalize structural regularities of the problem class.

The results presented in this work point toward a broader research direction beyond graph coloring: the development of neural solvers designed to operate in regimes close to algorithmic phase transitions. Many hard combinatorial optimization and inference problems, including SAT, Max-Cut, and error-correcting codes, exhibit structural transitions separating relatively easy from algorithmically challenging regimes.
While the present study is restricted to graph coloring, our results suggest that combining learned local update rules with problem-specific structural guidance and iterative search dynamics may provide a strategy for tackling similarly challenging problems. Extending the framework to different CSPs would, however, require problem-specific representations, differentiable constraint-violation objectives, and known solutions. Establishing whether the same advantages persist beyond graph coloring therefore remains an important direction for future work.

From an applied standpoint, scalable neural solvers for large combinatorial instances could impact domains such as communication networks, distributed computing, logistics, circuit design, and scheduling, where large-scale graph-structured constraints are ubiquitous. 

Several open directions emerge from this study. First, a deeper understanding of why the relationship between quadratic iteration scaling and the observed performance near the dynamical and Kesten–Stigum thresholds remains to be understood theoretically. Second, extending this framework to heterogeneous graph ensembles, weighted constraints, or real-world network topologies would test its robustness beyond random models. Third, integrating learned neural dynamics with classical message-passing or local-search algorithms could yield adaptive hybrid solvers that combine interpretability with learned flexibility.
Finally, an important direction is to reduce the reliance on labeled solutions during training, which currently represents a limitation of the proposed framework.

Ultimately, this work contributes to a growing perspective in machine learning: that learning can complement complexity theory and statistical physics in exploring the algorithmic limits of hard problems. By explicitly targeting phase boundaries and structural bottlenecks, neural solvers may help bridge the gap between theoretical optimality and practical scalability in combinatorial optimization and inference.
\section{Methods}\label{sec:methods}
In this section, we describe the model architecture and dataset. The loss function and training method are then introduced. Finally, the algorithm used for searching the solution is presented and motivated.
\subsection{Model}
We are interested in learning a map $f(\cdot;\mathcal{G})$ from a feature space $
\mathcal{X}\in\mathbb{R}^{k \times N}$ to a color space $\{1, \dddot{}, q\}^N$ such that:
\begin{equation}
    H_\text{Potts}\big[f(\{\mathbf{x}_i\}_{i\in\mathcal{V}};\mathcal{G})\big]=0
\end{equation} 
In our setting, each node is represented by a feature vector of dimension $k=q+1$, where the first $q$ components encode the color information and the last component encodes the node degree.
We choose to parameterize this map with a GNN model, implementing the transformation $f_{\boldsymbol{\theta}}(\cdot;\mathcal{G})$.
GNNs are designed to leverage the invariance properties of graph structures, making them effective for learning representations that respect the underlying graph topology.

In recent years, a plethora of architectures have been presented in the context of GNNs \cite{zhou2021graphneuralnetworksreview}. For our purposes, we have focused on one of the most general and expressive models, described in \cite{battaglia2018relationalinductivebiasesdeep}, and already used with promising results in solving various tasks \cite{TANG2023102616, WIEDER20201, wang2019graphnetspartialcharge}.
The model used in this work is based on a message passing network composed by $L$ layers. At each layer nodes features are updated as follows: at first for each edge an ingoing and an outgoing message are computed, (each node sends and receives a message from its neighbors) using the formula
\begin{equation}
    \textbf{m}_{ij}^{(l)} = \phi^{(l)}_{\theta_1^{(l)}} \left( \mathbf{x}_i^{(l-1)}, \mathbf{x}_j^{(l-1)} \right)\;,
\end{equation}
where $\textbf{m}_{ij}^{(l)}$ represents the message sent by the $j^\text{th}$ node and gathered by the $i^\text{th}$ node in the $l^\text{th}$ layer, while $\mathbf{x}_i^{(l-1)}$ and $\mathbf{x}_j^{(l-1)}$ represent the node features at the previous layer.
Next, the messages are aggregated using any permutation invariant function (e.g. sum, average, max). Together with the node features from the previous step \( \mathbf{x}_i^{(l-1)} \), the aggregated messages are used to update the node features:
\begin{equation}
    \mathbf{x}_i^{(l)} = \gamma^{(l)}_{\theta_2^{(l)}} \left( \mathbf{x}_i^{(l-1)}, \bigoplus_{j \in \mathcal{N}(i)} \textbf{m}_{ij}^{(l)} \right) \;,
\end{equation}
where $\mathcal{N}(i)$ is the set of neighbors of node $i$ in the graph.
The functions $\phi^{(l)}_{\theta_1^{(l)}}$ and $\gamma^{(l)}_{\theta_2^{(l)}}$, used to compute the messages and the node features respectively, are parameterized by two multi layer perceptrons (MLP). 
After the $L^\text{th}$ layer, all hidden and input features are passed to a third MLP $\Gamma_{\theta_3}$ to compute the final output of the node:
\begin{equation}
    \mathbf{x}_i^\text{output}=\text{softmax}\Big[\Gamma_{\theta_3}(\{\mathbf{x}^{(l)}_i\}_{l=0,\dots,L})\Big]
\end{equation}
The softmax activation makes the output feature vector normalized. In this way the $a^\text{th}$ component of $\mathbf{x}_i^\text{output}$ can be interpreted as a node's probability of having the $a^\text{th}$ color.
We denote the overall function implemented by the parametric model with $f_{\boldsymbol{\theta}}\big(\{\mathbf{x}_i^{(0)}\}_{i\in\mathcal{V}}\big)$, where $\boldsymbol{\theta}$ represents the set of all trainable parameters and we denote the input feature vector associated to the $i^\text{th}$ node with $\mathbf{x}_i^{(0)}$.
We highlight that by construction, the hidden features \( \mathbf{x}_i^{(l)} \) contain information about all neighbors at distance \(l\) from the $i^\text{th}$ node. Moreover the output function $\Gamma_{\theta_3}$ is able to use all intermediate features at once, effectively learning to process messages gathered at different graph distances.

\subsection{Dataset}\label{par:data}

We are interested in training our algorithm using both supervised and unsupervised loss terms. To do so, we need a dataset of graphs satisfying the GCP.
It is already well known that finding a good strategy to train neural networks to solve problems for which it is hard to have a solution is a challenging point. In this paper we overcome this problem, using the efficient planting procedure to obtain a random graph together with a solution of the GCP that could be however hard to find even for very smart algorithms \cite{krzakala2009hiding}. The ensemble of graphs generated with this planting procedure is usually referred to as the planted ensemble and, in some region of the parameters, is contiguous to the random Erdős–Rényi ensemble of graphs.

\subsection{Training and loss function}

In this work, we are interested in training a model that, given a colored graph with conflicting edges, produces a perfect coloring. Hence we are interested in finding a map
\begin{equation}\label{eq:map}
    f: \{1, \dddot{}, q\}^N \longrightarrow \{1, \dddot{}, q\}^N
\end{equation}
To this end, the model has been trained with a semi-supervised loss function, including an energy term and a term which measures the closeness of the output with the perfectly colored solution. The training is performed using a gradient-descent-based algorithm, therefore the loss function must be differentiable with respect to the model's parameters. This is not possible by using the Potts energy directly, as it is not differentiable. For this reason, regarding the energy term in the loss function, we have employed a continuous version of the Potts energy, introduced in \cite{schuetz2022graph}:
\begin{equation}
    H_\text{Potts}^c\big[\{\mathbf{y}_i\}_{i\in\mathcal{V}}\big] = \sum_{(i,j)\in\mathcal{E}}\langle \textbf{y}_i, \mathbf{y}_j\rangle \;,
\end{equation}
where the $a^\text{th}$ component of $\mathbf{y}_i$, indicated as $y_i^a$, is the probability of the $a^\text{th}$ color to be assigned to the $ i^\text{th}$ node. For this reason, both spaces described in Eq.~(\ref{eq:map}) are extended to $\mathbb{R}^{q\times N}$.

Regarding the supervised term in the loss function, this is inspired by a denoising autoencoder \cite{vincent2008extracting}, which forces the model to map corrupted inputs by adding noise (thus moving them away from the solution clusters) into inputs belonging to the solution manifold. In our case, the model is trained by taking a planted solution $\{\boldsymbol{\xi}_i\}_{i\in\mathcal{V}}$, written in the extended space, to which noise $\{\boldsymbol{\epsilon}_i\}_{i\in\mathcal{V}} \sim \mathcal{N}(\boldsymbol{0}, \hat{I})$ is added
\begin{equation}\label{eq:noise}
    \{\tilde{\mathbf{x}}_i\}_{i\in\mathcal{V}} = \sqrt{\alpha} \{\boldsymbol{\xi}_i\}_{i\in\mathcal{V}} + \sqrt{1 - \alpha} \{\boldsymbol{\epsilon}_i\}_{i\in\mathcal{V}}
\end{equation}
and including in the loss function a term which measures the closeness of the input planted graph to the output of the model.
At each training iteration, $\alpha$ is sampled from the uniform distribution $\mathrm{Uniform}(\alpha_{\min},\alpha_{\max})$.

A third contribution to the loss is given by an entropy term, measuring the average output node entropy. This term is used only during the initial steps of training to help the algorithm escape the region of parameter space that corresponds to the paramagnetic state $\Big(\big[f_{\boldsymbol{\theta}} (\{\textbf{x}_i\}_{i\in\mathcal{V}})\big]_j^a = \frac{1}{q}, \forall j \in \mathcal{V}, \forall a \in \{1,\dddot{},q\}\Big)$. We have empirically observed that this term does not lead to a final lower energy but increases the convergence speed.
The final loss function is then given by
\begin{multline*}
    L(\{\textbf{x}_i\}_{i\in\mathcal{V}}, \{\boldsymbol{\xi}_i\}_{i\in\mathcal{V}}; \boldsymbol{\theta}) = \\
    \mathbb{E}_{\{\boldsymbol{x}_i\}_{i\in\mathcal{V}}} \big[ h \left[ f_{\boldsymbol{\theta}} (\{\textbf{x}_i\}_{i\in\mathcal{V}}) \right] \big] \\
    + \eta_1\, \mathbb{E}_{\{\boldsymbol{x}_i\}_{i\in\mathcal{V}}} \big[ S \left[ f_{\boldsymbol{\theta}} (\{\textbf{x}_i\}_{i\in\mathcal{V}}) \right] \big] \\
    - \eta_2\, \mathbb{E}_{\{\boldsymbol{x}_i\}_{i\in\mathcal{V}}} \big[ O \left[ f_{\boldsymbol{\theta}} (\{\textbf{x}_i\}_{i\in\mathcal{V}}), \{\boldsymbol{\xi}_i\}_{i\in\mathcal{V}} \right] \big]
\end{multline*}
where $\eta_1$ and $\eta_2$ are relative weights and the three contributions are respectively an \textit{energy term}, namely the percentage of conflicting edges, the \textit{features entropy} and the \textit{overlap} with the planted solution. These terms are given by
\begin{align*}
&h \left[ f_{\boldsymbol{\theta}} (\{\textbf{x}_i\}_{i\in\mathcal{V}}) \right] \notag = \frac1M \sum_{(i,j)\in\mathcal{E}}\langle f_{\boldsymbol{\theta}} (\{\boldsymbol{x}\})_i, f_{\boldsymbol{\theta}} (\{\boldsymbol{x}\})_j\rangle\\
&S \left[ f_{\boldsymbol{\theta}} (\{\textbf{x}_i\}_{i\in\mathcal{V}}) \right] = -\frac{1}{N}\sum_{i\in\mathcal{V}}f_{\boldsymbol{\theta}} (\{\boldsymbol{x}\})_i\cdot \log_2 f_{\boldsymbol{\theta}} (\{\boldsymbol{x}\})_i\\
&O \left[ f_{\boldsymbol{\theta}} (\{\textbf{x}_i\}_{i\in\mathcal{V}}), \boldsymbol{\xi}_i \right] = \frac{1}{|\mathcal{V}|}\sum_{i\in\mathcal{V}}\langle f_{\boldsymbol{\theta}} (\{\boldsymbol{x}\})_i, \boldsymbol{\xi}_i\rangle
\end{align*}
where $A_{ij}$ is the adjacency matrix and 
\begin{equation}
    h \left[\{\textbf{x}_i\}_{i\in\mathcal{V}} \right] = \frac{1}{M}H_\text{Potts}^c\big[\{\mathbf{x}_i\}_{i\in\mathcal{V}}\big]
\end{equation}
The overlap term is necessary to break the permutation symmetry of the colors. In fact, the energy and entropy terms in the loss are invariant for color permutations. Consequently the loss landscape contains $q!$ identical regions, each corresponding to equivalent, permuted colorings. Breaking this symmetry simplifies the map that the model needs to learn, as the loss landscape is modified so that only the planted solution has the lowest loss, guiding the optimizer more efficiently.

The training phase of our model is detailed in Algorithm A.2 in the Supplementary Information.

\subsection{Coloring}\label{subsec:coloring}
In this paragraph, we detail the coloring process using our trained model $f_{\boldsymbol{\theta}}\big(\cdot\big)$.
For a sufficiently expressive model, we would expect to find the perfect coloring with a single forward pass ($N_\text{iter}=1$).
For our model, however, we have empirically observed that this is not the case, as the output energy is further reduced with subsequent applications of the same trained model.
We call $N_\text{iter}$ the number of iterations, that is, subsequent applications of the model.
Additionally, we have observed that the algorithm used is characterized by fixed points that do not correspond to the solutions of the GCP. This causes the algorithm to get stuck in sub-optimal configurations. We have verified that performances are greatly improved by adding noise after each forward step, as described in Eq.~\ref{eq:noise}.
We experimented with different choices for noise scheduling.
During this noise-annealing procedure, the neural network gradually reduces the energy of the input graph. If it is performed for a large enough number of iterations, this procedure eventually leads the graph into the attraction basin of a zero-energy solution. The scheduling of noise from large to smaller values has the meaning of letting the model explore a larger portion of the color space in the initial steps, while making the model collapse towards the zero-energy solution in the final steps. We have empirically observed how this procedure produces very similar results even when a solution is not guaranteed to exist.
A more detailed discussion of how noise modifies the forward step, together with a detailed description of the coloring procedure, is provided Section A.3 of the Supplementary Information.

\subsection{Software and Computational Tools}
Graph generation and data preprocessing were performed using custom Python and C implementations. The graph neural network was developed and trained using Python 3.9.25, PyTorch 2.8.0, PyTorch Geometric 2.6.1, PyTorch Lightning 2.5.1, and torch-scatter 2.1.2. Additional numerical computations were performed using NumPy 1.26.4. Experimental results were analyzed and visualized using custom Python scripts and Matplotlib 3.9.4. 

\section*{Data availability}
The data supporting the findings of this study, including the numerical data underlying the figures, are available on Figshare at https://doi.org/10.6084/m9.figshare.33129692. 

\section*{Code availability}
The code used to generate the datasets and implement the models is publicly available on GitHub and archived on Zenodo \cite{graphcoloring_code}.

\section*{Author Contributions}
L.C.,A.C.,S.G., developed the study design and the methodological framework. L.C.,A.C. curated the codes, conducted models training, generated predictions, and performed the analyses.  F.S. performed tests. M.C.A. curated the generation of synthetic data.
S.G.,F.R.-T.,M.C.A., supervised the methodological framework.  M.C.A.,F.R.-T., provided statistical physics expertise. L.C.,A.C.,S.G., wrote the draft of the manuscript. All authors discussed the results and improved the writing of the paper.

\section*{Funding Statement}
This work was supported by PNRR MUR project PE0000013-FAIR and by the “National Centre for HPC, Big Data and Quantum Computing”, Project CN\_00000013, CUP B83C22002940006, NRRP Mission 4 Component 2 Investment 1.4,  Funded by the European Union - NextGenerationEU.

\section*{Competing Interests Statement}
The authors declare no competing interests.

\section*{Figure Legends}

\noindent
\textbf{Fig. 1 } Schematic illustration of the evolution of the solution-space landscape
with increasing average connectivity $c$. The red cross represent the
planted solution. At low connectivity, the planted solution belongs to a
large connected region of solutions. Above the dynamical transition $c_d$, the solution space fragments into an exponential number of clusters. At the condensation transition $c_c$, the solution measure concentrates onto a subexponential number of dominant clusters, one of which contains the planted solution. The satisfiability threshold $c_s$ marks the transition beyond which typical random instances become unsatisfiable. The Kesten--Stigum threshold $c_{\mathrm{KS}}$ marks the detectability of the planted structure. Blue regions denote solutions, whereas orange regions represent local minima.

\vspace{1em}

\noindent
\textbf{Fig. 2 }
(DTR Model) Extensive energy as a function of the number of iterations for four representative connectivity values, $c = 12.0, 13.0, 13.5, 14.0$. Each data point is obtained by averaging results over 50 graphs with $N=$10k nodes. Error bars indicate one standard deviation across graph instances. Circles indicate values for planted graphs, while crosses indicate values for random graphs. Error bars indicate standard error of the mean
across graph instances.
\vspace{1em}

\noindent
\textbf{Fig. 3 }(DTR Model) \textbf{Main panel} Mean intensive energy as a function of connectivity, for both random and planted graphs, and a number of nodes varying from $N=2500$ to $N=50$k. Each data point is obtained by running the algorithm for $N_\text{iter}=200$k iterations and averaging the energy over 50 graphs. Error bars indicate one standard deviation across graph instances. \textbf{Inset} The mean extensive energy grows with the size at each value of $c$ and does not allow to determine the algorithmic threshold in the large $N$ limit. Error bars indicate the standard error of the mean.

\vspace{1em}

\noindent
\textbf{Fig. 4 }
(DTR Model) Probability of finding a solution as a function of the graph connectivity $c$, for sizes $N\in[1024, 2048, 4096, 8192]$ and a number of iterations scaling as $N_\text{iter} = 0.04\,N^2$. Each point is the average over 400 graphs. Error bars indicate 95\% Wilson binomial confidence intervals.
\vspace{1em}

\noindent
\textbf{Fig. 5 }(DTR Model) Comparison of the probability of finding a solution as a function of the graph connectivity $c$ for different solving algorithms. Values are obtained by averaging over $400$ graphs of $N=1024$ nodes. The number of iterations $N_\text{iter}$ is chosen to ensure the same wall-clock execution time across all algorithms. The tested algorithms are a recurrent Physics-Inspired GNN (rPI) \cite{schuetz2022graph}, Belief Propagation (BP) \cite{Skenderi2026}, Simulated Annealing (SA) \cite{angelini2025algorithmic}, our graph neural network (GNN), and Focused Metropolis Search (FMS) \cite{Seitz_2005}.  Error bars indicate 95\% Wilson binomial confidence intervals.

\vspace{1em}

\noindent
\textbf{Fig. 6 }
(KSR Model) \textbf{Main panel} Probability of finding a solution as a function of the graph connectivity $c$, for sizes $1000\le N\le 20000$ and a number of iterations scaling as $N_\text{iter} = 10^{-4}\,N^2$. Each point is the average over $2\,10^6/N$ graphs. \textbf{Inset} The mean overlap with the planted coloring of the configuration output by the algorithm. Error bars indicate 95\% Wilson binomial confidence intervals.

\end{document}